\documentclass[final,3p,times]{elsarticle}

\usepackage{amssymb}
\usepackage{amsmath}
\usepackage{xcolor}
\usepackage{caption}
\usepackage{subcaption}
\usepackage{graphicx}
\usepackage{algorithm2e}
\usepackage{float}
\usepackage{natbib}

\newcommand{\figref}[1]{Fig.~\ref{fig:#1}}
\newcommand{\secref}[1]{Section~\ref{sec:#1}}
\newcommand{\tblref}[1]{Table~\ref{tbl:#1}}
\newcommand{\algoref}[1]{Algorithm~\ref{algo:#1}}

\newcommand{\eqnref}[1]{Equation~\ref{eq:#1}}

\usepackage{lineno}
\begin{document}
\begin{frontmatter}



\newcommand{\aimi}{AIMI}

\title{\aimi: Leveraging Future Knowledge and Personalization in Sparse Event Forecasting for Treatment Adherence}

\author[scai,chs]{Abdullah Mamun\corref{cor1}} 
\cortext[cor1]{Corresponding author. Email: a.mamun@asu.edu.}
\author[wsu]{Diane J. Cook} 
\author[chs]{Hassan Ghasemzadeh} 
\address[scai]{School of Computing and Augmented Intelligence, Arizona State University, Phoenix, AZ 85054, USA}
\address[chs]{College of Health Solutions, Arizona State University, Phoenix, AZ 85054, USA}
\address[wsu]{School of Electrical Engineering and Computer Science, Washington State University, Pullman, WA 99164, USA}

\begin{abstract}
Adherence to prescribed treatments is crucial for individuals with chronic conditions to avoid costly or adverse health outcomes. For certain patient groups, intensive lifestyle interventions are vital for enhancing medication adherence. Accurate forecasting of treatment adherence can open pathways to developing an on-demand intervention tool, enabling timely and personalized support. With the increasing popularity of smartphones and wearables, it is now easier than ever to develop and deploy smart activity monitoring systems. However, effective forecasting systems for treatment adherence based on wearable sensors are still not widely available. We close this gap by proposing \textbf{A}dherence Forecasting and \textbf{I}ntervention with \textbf{M}achine \textbf{I}ntelligence (AIMI). AIMI is a knowledge-guided adherence forecasting system that leverages smartphone sensors and previous medication history to estimate the likelihood of forgetting to take a prescribed medication. A user study was conducted with 27 participants who took daily medications to manage their cardiovascular diseases. We designed and developed CNN and LSTM-based forecasting models with various combinations of input features and found that LSTM models can forecast medication adherence with an accuracy of 0.932 and an F-1 score of 0.936. Moreover, through a series of ablation studies involving convolutional and recurrent neural network architectures, we demonstrate that leveraging known knowledge about future and personalized training enhances the accuracy of medication adherence forecasting. Code available: https://github.com/ab9mamun/AIMI.
\end{abstract}



\end{frontmatter}
\newcommand{\aimi}{AIMI}
\section{Introduction}
Approximately 6.7 million adults are affected by heart failure (HF) in the U.S. \cite{martin20242024}. The number of American adults with HF has increased by 31.4\% in the last 10 years \cite{go2014heart}. It is a deadly disease with a 5-year mortality rate of 50\% and a 10-year excess mortality rate of 36-40\% \cite{taylor2019trends, drozd2021association}. Medication adherence is crucial for many HF patients as well as individuals suffering from hypertension, hyperlipidemia, and diabetes \cite{qin2023evidence, hood2018association, fitzgerald2011impact, brown2011medication}. Despite the risks, people often fail to adhere to the prescribed medications \cite{mykyta2023characteristics, kleinsinger2018unmet}. Lifestyle interventions, such as smartphone apps and reminders, can improve medication adherence, but low-cost reminder solutions are not as effective as intensive interventions in increasing medication adherence \cite{de2023effect, kini2018interventions, choudhry2017effect, kini2018interventions}. 

In this study, we aim to design an effective intervention system by accurately forecasting whether a person is likely to miss their prescribed medication. Treatment adherence forecasting was addressed in different prior publications in different application domains. An adherence prediction system with Long Short-Term Memory (LSTM) networks was developed in \cite{gu2021predicting} that can predict adherence to specific injection-based treatments with 77.35\% accuracy. In that study, Internet-connected sharps bins were used to collect treatment events. Another study developed adherence forecasting among individuals with psychiatric disorders with an AUC of 0.87 \cite{koesmahargyo2020accuracy}. Several publications from the past addressed the forecasting of health events such as hospital readmission, development of heart failure, and other adverse health outcomes \cite{choi2017using, michailidis2022forecasting, vieira2023forecasting}. Disease progression and prediction of health outcomes were previously addressed using electronic health records (EHR) \cite{alaa2019attentive}. LSTM-based models were proposed for forecasting daily step counts \cite{mamun2022multimodal, mamun2024multimodal}. \cite{huang2020medical} forecasts the medical service demand with a hybrid model based on ARIMA and a self-adaptive filtering method. Using future knowledge to improve prediction is a relatively new idea, and prior work in this area is very limited \cite{qi2021known}. Healthcare datasets collected in the wild face difficulties related to data noise, missing data, and natural human variance \cite{marino2021missing}. In this case, any additional available information should be considered for inclusion in the feature set. However, current literature does not provide enough evidence to draw a reliable conclusion about the impact of factors such as time context, location, or future knowledge—i.e., known future events—on a person’s treatment adherence.

We hypothesize that medication adherence can be affected by a person's temporal and behavioral context, such as the day of the week, location, and activity. Consider the situation where someone is in the middle of a trip or went to bed later than usual, possibly because they are sleeping late the next morning. In these two cases, they may not take the medication at the prescribed time. Although adherence forecasting using past electronic health records has been explored in the past \cite{kumamaru2018using}, forecasting such events using sensor or physical activity data or future knowledge features did not get sufficient attention.

This research aims to bridge this gap by formulating a sparse event forecasting problem and exploring context-aware and knowledge-guided deep learning solutions. We propose a smart health system, \aimi  (\textbf{A}dherence Forecasting and \textbf{I}ntervention with \textbf{M}achine \textbf{I}ntelligence), that considers not only the current activity and history of adherence but also the time-context and known information of future prescriptions. To evaluate our method, we conducted a user study with N=27 participants who take medications to manage cardiovascular diseases. Then Convolutional Neural Networks (CNN) and LSTM models were trained and evaluated as potential forecasting models for the AIMI system.

Furthermore, we include an incremental learning algorithm for training the system in a resource-constrained training environment, e.g., limited memory. This incremental learning not only facilitates on-device learning in healthcare systems but also ensures the security of the data as it eliminates the need to transfer any data to third-party servers to train neural networks. In such training settings, the performance of neural networks is prone to worsen for specific instances after they are trained on new instances \cite{chen2020mitigating} which we overcome with personalization of the final model.

In summary, our main contributions are: i) investigating the challenges of predicting and forecasting treatment adherence, ii) proposing the AIMI system for forecasting treatment adherence using sensor data, adherence history, and \textit{future knowledge} features, iii) providing implementations of our proposed method using CNN and LSTM models and a systematic comparison of the models for this task, iv) conducting a user study and evaluating the AIMI on the collected dataset, v) proposing an incremental learning algorithm for resource-constrained environments, and vi) determining the impact of knowledge guidance and personalization of machine learning models for treatment adherence forecasting.

\section{Background and Related Work}
\subsection{Treatment Adherence and Personalization}
Treatment adherence can be defined in a few possible ways. One form of nonadherence is not taking the medication at the prescribed time. Whether a person will adhere to the prescribed medications can be predicted with baseline questionnaires \cite{mirzadeh2022use}. Another form of nonadherence can be violating a medical temporal constraint (MTC) associated with the medicine, such as a recommendation to take a particular medication after a meal or before a meal. \cite{seegmiller2023actsafe} predicts the violation of a medical temporal constraint with a behavioral prediction model, HERBERT, that utilizes a personalized training process.

Predicting or forecasting the health events of a person differs from the same statistical variables that are affected by the actions of multiple entities of the world. A generic classifier may achieve strong performance on one set of participants' data but suffer for other participants \cite{sah2022stressalyzer}. That is why fine-tuning the generic models is necessary for some participants for such challenging problems \cite{azghan2023personalized}. Prior work \cite{sah2022stressalyzer} demonstrates that the classification accuracies for an individual in a stress classification problem can be improved by a margin of up to 39\% after employing personalized training.

\subsection{Forecasting with Deep Learning}
Time-series forecasting is a complex task that has been around for a considerable length of time \cite{benidis2022deep}. In earlier work, a language model-based forecaster predicted world events including climate and geopolitical conflicts \cite{zou2022forecasting}. While transformers \cite{vaswani2017attention} have excelled in many different areas including natural language processing \cite{devlin2018bert, jiang2023mistral}, image and video data processing  \cite{caron2021emerging, islam2022long}, and language generation \cite{brown2020language, touvron2023llama, chowdhery2023palm}, they may not perform well for forecasting stationary time-series data. Non-stationary transformers can overcome this issue by re-incorporating nonstationarity on feed-forward layers or using De-stationary attention \cite{liu2022non}.

Recurrent networks have shown promise for event forecasting \cite{chaudhary2021jointly, zhang2024forecasting}.  Forecasting can be univariate or multivariate, depending on how many variables we want to jointly predict. The LSTNet model, a combination of a CNN, LSTM, and autoregressive layer-based neural network, has proven to be effective in multivariate forecasting \cite{lai2018modeling}. However, the LSTNet method scales poorly because the input layer's size depends on the number of variables (e.g. power generation in $x$ power plants) we want to forecast.

There are important questions to answer in the area of forecasting with future knowledge regarding their impact and effective ways to incorporate them. Qi et al. \cite{qi2021known} proposed a knowledge-guided transformer to predict the demand for products of Alibaba. Li et al. \cite{li2022kpgt} followed a similar process to predict molecular properties of matter. Additionally, Liu et al. \cite{liu2023generative} provided knowledge-guided decoding for academic knowledge graph completion.

\subsection{Classical Methods of Forecasting}
Before deep learning was established, various classical methods were proposed to solve natural and urban problems with forecasting. In particular, Gagliardi et al. \cite{gagliardi2017probabilistic} introduced a Markov chain model to forecast short-term water demand in different areas. Another popular forecasting technique is imitation learning, where models are trained to mimic an expert’s decisions \cite{minor2015data}.  Lee and Tong \cite{lee2011forecasting} proposed a forecasting algorithm based on ARIMA and genetic programming and validated their method using three different datasets. Often, deep recurrent models demonstrate superior performance over autoregressive models and linear regression models \cite{mamun2024multimodal, siami2018comparison} but in specific settings, ARIMA or seasonal ARIMA can still achieve better performance than LSTM or GRU models \cite{liu2021short}.

\subsection{Challenges of Time-Series Sensor Data}
Wearable sensor datasets in free-living environments are often accompanied by numerous challenges, such as low sampling rates, missing data or missing sensors, and noisy or inconsistent labels. Alternative imputation methods are available for recovering missing sensor data, such as CNN-based \cite{mamun2022designing}, GAN-based\cite{hussein2024sensorgan}, clustering-based \cite{Hussein2024}, etc. Another common challenge is the selection of channels. An efficient channel selection process was presented by \cite{sah2023stress}. One way to overcome the challenge of a low sampling rate is by extracting statistical features \cite{mirzadeh2019labelmerger}.

\section{Materials and Methods}
\subsection{Sparse Event Forecasting with Future Knowledge}
\label{sec:formulation}
Given a dataset, $\mathcal{D} = (S, E)$, containing wearable sensor readings, $S$, and sparse event data, $E$ for a particular participant, the goal is to forecast future values of $E$. The sensor and event data are collected over $\mathcal{T}$ seconds. The sparse event data were sampled at a much lower rate than the sensor data, i.e. $\nu_S >> \nu_E$, where $\nu$ represents the sampling frequency of a variable.

Therefore, for the same time interval, there will be more sensor data samples than sparse events.  Assuming there are $N$ data points of sensor data and $T$ data points of event data within the duration of  $T$, let us denote those data points as $S_{1..N} = (S_1, S_2, ..., S_N)$, and $E_{1..T} = (E_1, E_2, ..., E_T)$. Here, $E_i$ represents the action of taking medication at time step $i$, and it can take a value of either 0 or 1.

To accomplish the forecasting task, we assume that the value of $E$ at time point $T+L$ is correlated with the present and past values of $S$ and $E$. This can mean either i) the past $S$ and $E$ can partially influence the future values of $E$; or ii) the future value of $E$ is affected by an external factor that also affects the present and past values of $S$ and $E$. This forecasting problem can be written as the following equation if $f_1$ is the forecasting function.

\begin{equation}
\label{eq:first_forecaster}
    \hat{E}^{f_1}_{T+1} = f_1(S_{1..N} , E_{1..T})
\end{equation}

Consider a scenario in which we have additional knowledge about the future that is outside the dataset, $\mathcal{D}$. An example of a knowledge feature is the scheduled time for a meeting or a sales promotion. In the context of this paper, a knowledge feature is the prescribed time to take medication. Let us consider one knowledge parameter, $K$. One distinct property of this knowledge parameter is that it can be known in advance for even future time steps. Let $L$ be the number of time units into the future (time $T+L$) for which external knowledge is available at time $T$. That means, at time $T$, we can have the values for future knowledge features, $K_{T+1..T+L}$. The knowledge parameters can potentially be added as an extra parameter to the forecasting function of \eqnref{first_forecaster}. Then, the next evolution of the forecaster would be as follows:

\begin{equation}
\label{eq:second_forecaster}
    \hat{E}^{f_2}_{T+1} = f_2(S_{1..N} , E_{1..T}, K_{1..T+L})
\end{equation}

In \eqnref{second_forecaster}, we use knowledge parameters $K_{1..T+L}$, among which only the values for the future time steps, i.e. $K_{T+1, .., T+L}$ will be considered \textit{future knowledge}, whereas, $K_{1..T}$ will be considered present and past knowledge. The overall coverage of this paper will be to find out how future knowledge can improve the performance of a forecaster when integrated into the feature set of a neural network. Furthermore, this work also investigates i) whether the performance is negatively affected if the low-sampled feature set is removed from \eqnref{first_forecaster}, and ii) whether the performance drop can be recovered using future knowledge. These scenarios are presented in \eqnref{third_forecaster} and \eqnref{fourth_forecaster}.
\begin{equation}
\label{eq:third_forecaster}
    \hat{E}^{f_3}_{T+1} = f_3(S_{1..N})
\end{equation}

\begin{equation}
\label{eq:fourth_forecaster}
    \hat{E}^{f_4}_{T+1} = f_4(S_{1..N}, K_{T+1..T+L})
\end{equation}

Note that in \eqnref{fourth_forecaster}, $E$ is not present in the feature set. However, only $K$'s future time steps, or $K_{T+1..T+L}$, are used here. This formulation ignores past knowledge features as well.

\begin{figure}[!t]
\centerline{\includegraphics[width=0.6\columnwidth]{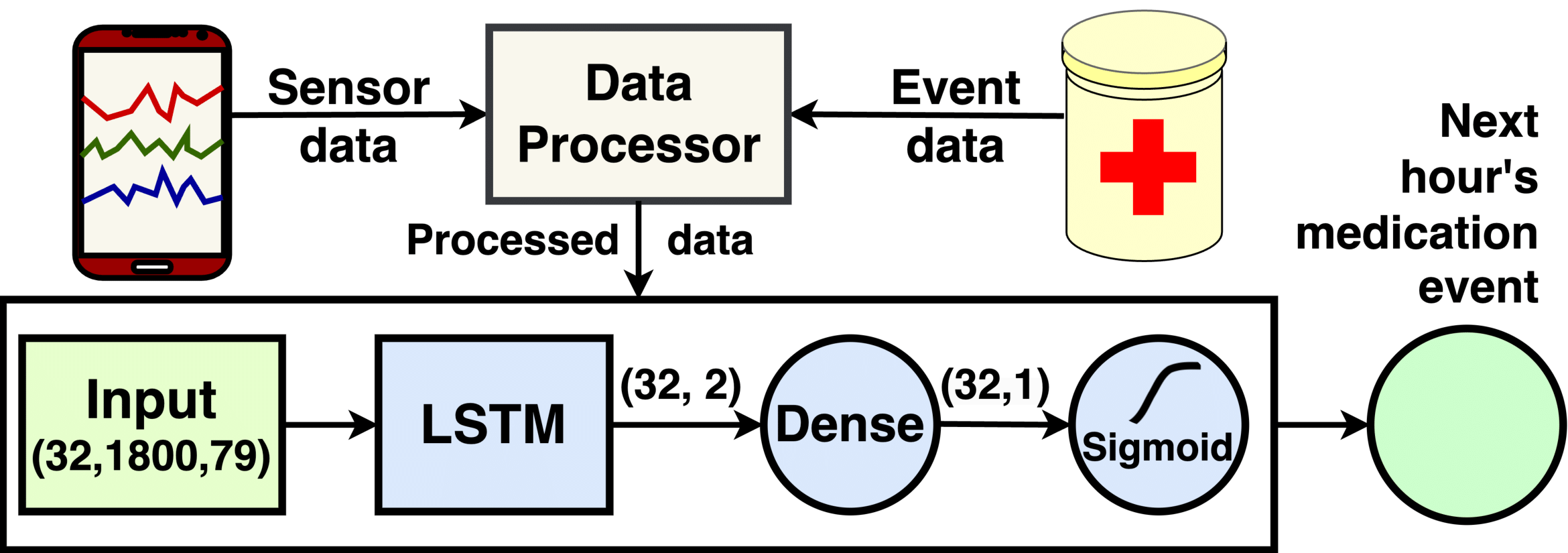}}
\caption{An overview of the AIMI system with the architecture of the LSTM-based forecasting model. The shapes of the data for a batch size 32 and forecasting with sensor, event, and knowledge features are indicated in parentheses.}
\label{fig:lstm}
\end{figure}

We implement this solution with a machine learning model and evaluate the model on a medication adherence dataset. Each forecasting function is a combination of data processing and a machine learning model, $M$. The sensor data, $S$, and the event data, $E$, may not be in the usable shape for the model, $M$. Therefore, the forecasting function will process the data by cleaning it, creating features,  addressing imbalances, and handling related issues. Data processing details are discussed in \secref{processing}. The architecture of the LSTM model is displayed in \figref{lstm}, and readers are referred to \ref{sec:app1} for the details of the CNN model used for the experiments.

\subsection{Dataset}
To validate the method described in \secref{formulation}, a medication adherence dataset is chosen \cite{mirzadeh2022use}. This dataset is derived from a clinical study that was conducted with people who are at risk for atherosclerotic cardiovascular disease and take medications for their condition. The study was approved by an Institutional Review Board (IRB) and all participants provided informed consent.

Study participants performed their regular activities while their smartphone sensors continuously recorded data that reflected their activities. The details of the data used for this study are described in \secref{processing} and \tblref{feature_summary}. Medication events were collected from electronic pillboxes. The sensor data are sampled at 1Hz. For most participants, the medication events are collected at 1 event/day, though for some participants, these were collected at 2 events/day, as prescribed by their physicians. Initially, 27 participants were available to form the dataset for this paper. Among them, data from 2 participants were removed due to unresolvable collection errors. From the remaining 25 participants, randomly 3 participants' data are held out for future testing purposes. Thus, we used 22 participants' data for training and testing the models used in the experiments.

For each participant, the sensor data and the event data are saved in separate files. A sensor data file has sensor data sampled at a rate of 1Hz including yaw, pitch, roll, 3-axial acceleration, 3-axial rotation, location coordinates, altitude, horizontal and vertical accuracies, and speed features. However, there was a 5-second pause after every 5 seconds of data collection. Therefore, 5 sensor readings were collected every 10 seconds. An event data file contains the medication event data for each day for a person. If the person takes a medication, it is recorded with the timestamp. If no record is found for a day, it implies that the person did not take medication on that day. Moreover, the prescribed times are also present for those days. For some participants, more than one medication was prescribed for each day but for other participants, only one medication was prescribed. Therefore, usually, the medication events are sampled at 1-2 samples/day frequency.

\subsection{Data Preprocessing}
\label{sec:processing}
 To make these data usable for neural network training, the sensor data and the event data are first combined to create a merged data file. This merged data file contains sensor data for each second, the previous medication event, and the next medication event. The timestamps of the previous and next medication events were included, as well as the previous and next prescribed times. The merged data file is then processed to add more features, such as contextual features (day of the week, hour of the day) and additional derived features. For example, whether any medication was taken in the last 2, 3, 6, 12, or 24 hours and whether any medication is taken the next hour. The medication event of the next hour is chosen as the target variable that we want our neural network model to forecast.

 To create training data points, a sliding window was slid over the time series data with a duration of 3600 seconds and 50\% overlap. For each sensor data entry, the reading was recorded with both local and UTC timestamps. These were converted to relative timestamps by subtracting the current timestamp from the timestamp of the prescribed time for the next medication event. Relative timestamps were calculated using prescribed time, which is \textit{future knowledge}. Hence, the relative timestamps were considered future knowledge features. In the dataset, sensor data were collected at 1Hz frequency with a 5-second pause after every 5 seconds of data collection. In other words, 5 sensor readings were collected every 10 seconds on average. The length of the sliding windows was set to 1800 samples, which is 3600 seconds of data.
 
\begin{figure}[!t]
\centerline{\includegraphics[width=0.55\textwidth]{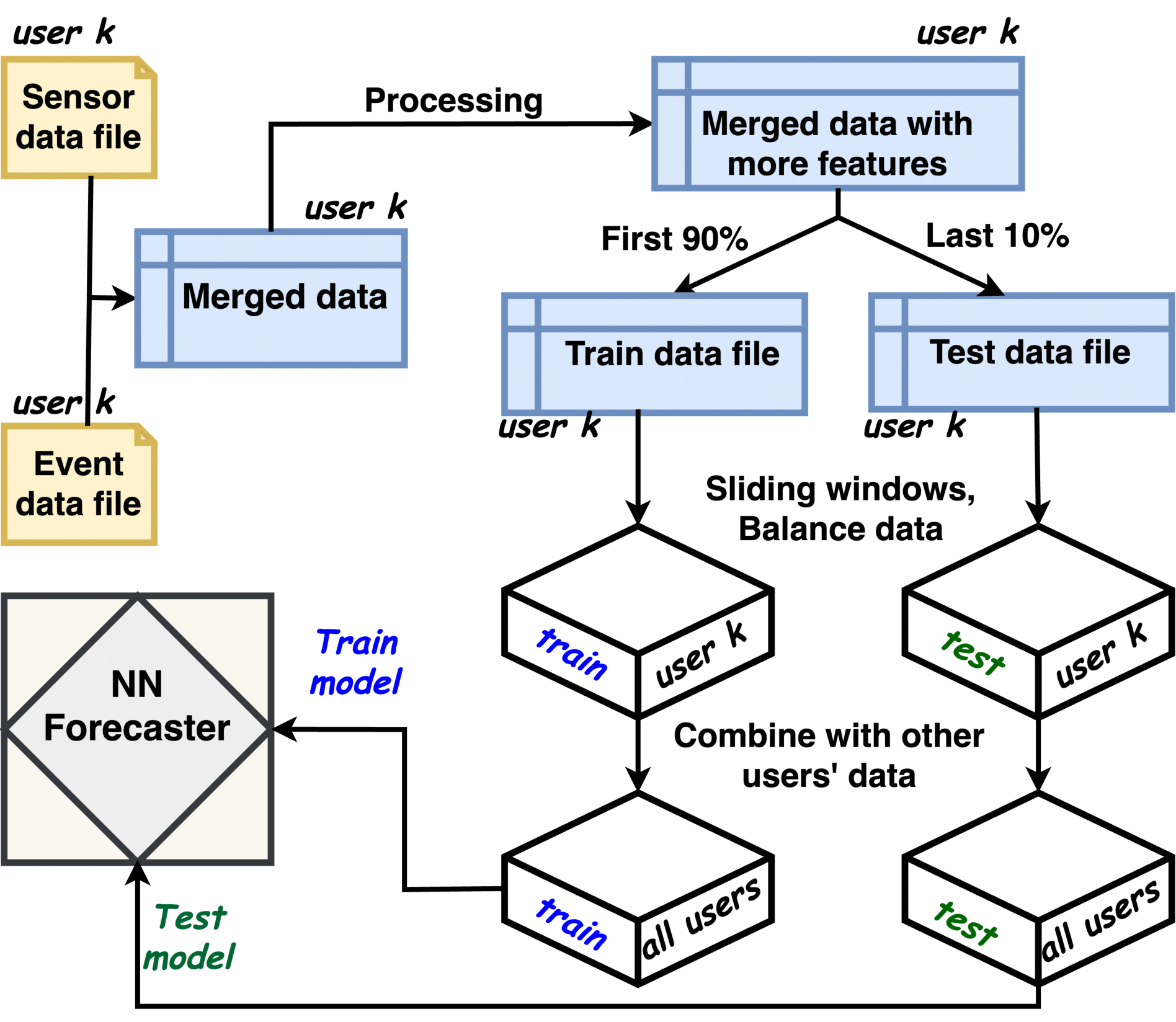}}
\caption{An illustration of the AIMI system's data processing method and separation of training and test data.}
\label{fig:dataflow}
\end{figure}

 The prescribed times are the same for a patient across different days. Whether the person takes the medication, `medication next hour', may depend on the `next prescribed hour' but not the other way around. Thus, these relative timestamps got the information regarding the \textit{future knowledge}, prescribed time, and had to be chosen as future knowledge features. For the experiments conducted, the final feature set that was used is presented in \tblref{feature_summary}. A logical flow of the process is illustrated in \figref{dataflow}.

\begin{table}[htbp]
\centering
\caption{Dataset features. Here, the \textit{Relative timestamp} and \textit{Next prescribed hour} features embed knowledge of the future. \textit{Medication next hour} is the forecast variable. Type N=numeric, B=boolean.}
\begin{tabular}{lll}
\hline
\textit{Feature}   &  Type &  \textit{Dimension} \\
\hline
\textbf{High-resolution features (H)} &         & \textbf{14} \\
Yaw, Pitch, Roll            & N            & 3   \\
Rotation rate (x,y,z)    & N  & 3   \\
Acceleration (x,y,z)     & N  & 3  \\
Latitude, longitude  & N & 2 \\
Altitude, horizontal accuracy, speed  & N     & 3   \\
\hline
\textbf{Low-resolution features (L)} & &  \textbf{39} \\
Last medication event & B        & 1  \\
Last prescribed time   & N      & 1 \\
Last event hour & N & 1 \\
Day of the week  & B & 7 \\
Hour of the day  & B &  24 \\
Medication events of  &  &  \\
last 2, 3, 6, 12, and 24 hours & B & 5\\
\hline
\textbf{Future knowledge features (K)} &         & \textbf{26} \\
\textit{Relative timestamp}      & N                 & \textit{2}   \\
\textit{Next prescribed hour} & B & \textit{24} \\
\hline
\textbf{Medication next hour (target)} & B & \textbf{1} \\
\hline
\end{tabular}
\label{tbl:feature_summary}
\end{table}

\subsection{Training Process}
\RestyleAlgo{ruled}
\begin{algorithm}[b]
\caption{Incremental training process for a large dataset on a machine with limited memory.}
\label{algo:incrementaltraining}
     Find size $S$ for the data, that the machine can handle in one training session\;
     Divide the training data, $D_{train}$ into $n$ disjoint chunks, $Chunk[1]...Chunk[n]$ so that $D_{train} = Chunk[1] \cup Chunk[2] \cup ... \cup Chunk[n]$ and $Size(Chunk[i]) \leq S$ for  $1 \leq i \leq n $\; 
     $i \gets 1$\;
    \While{$i \leq n$}{
         Create and compile model $M$\;
         \If{$i \geq 2$}{
            Load weights from file $model\_weights$ to model $M$\;
         }
         Train model $M$ with $Chunk[i]$\;
         Save weights of model $M$ to file $model\_weights$\;
         $i \gets i+1$\;
    }
\end{algorithm}
During the initial experiments for model choosing and the first trial, the models were trained on an Intel(R) Core(TM) i7-7500 CPU with a clock speed of 2.7 GHz and 16 GB RAM (random access memory). Then the LSTM experiments were repeated on a computation node of a supercomputer with 32 cores, 64 GB RAM, and an NVIDIA A100 GPU. The models were trained in an incremental manner to keep the training pipeline compatible with training environments with limited computing resources. We used total data from 22 participants to train the models. The largest data file of the ``Merged data with more features'' step, as shown in \figref{dataflow}, for a single participant was around 1.18 gigabytes (GB), whereas the total RAM of our CPU-based training machine was 16 GB. Hence, the training was done in 6 sessions to cover all 22 participants. At first, the 22 participant IDs were shuffled and then separated for 6 different sessions. Those sessions used 1, 4, 4, 4, 5, and 5 participants' \textit{training data} for training and the same 1, 4, 4, 4, 4, and 5 participants' \textit{test data} for testing, respectively. Each session saved the model weights after the training was complete and the next session (if any) loaded that weight before starting the training. Thus, after each training session, the cumulative number of participants' training data that was used would be 1, 5, 9, 13, 17, and 22, respectively. This incremental training is further described in \algoref{incrementaltraining}. The whole process was repeated 3 times with three different random orders of the participants.

\subsection{Evaluation Method}
In this work, training data and test data were separated in chronological order for each of the 22 participants, as shown in \figref{dataflow}. Accuracy and macro average precision, recall, and F-1 scores were chosen for the evaluation of the performance. The training and test sets were divided before shuffling or balancing the data. ADASYN was used for balancing the data by the generation of synthetic data points near the existing data points of the minority class \cite{he2008adasyn}. The training set and the test set were processed and balanced independently but through similar processes. This ensures that there is no data leakage between the training and the test set. Moreover, the feature sets were carefully considered so that any information related to the target forecast variable, medication event of the next hour, could not influence the feature variables.

\section{Results}
\label{sec:results}
\subsection{Comparison Between CNN and LSTM}
The main goal of this study is to present the importance of using future knowledge in the feature set. To accomplish this, we test two candidate models and proceed with the one that performs better in the first round of experiments. We have tested our method with LSTM and CNN models. The initial analysis found that LSTM provided overall higher and more stable performance in forecasting. Moreover, the performance of CNN did not improve much with the increasing number of epochs.

We decided that both CNN and LSTM models would be trained for approximately the same duration. The total duration of training was tuned by setting the number of epochs. It was observed that the LSTM model needed about 10 epochs to reach an accuracy and F-1 score close to 90\% when trained on 22 participants' data with all features. The designed CNN model had 189,972 parameters compared to the 651 parameters of the LSTM model. However, the CNN model trained faster per epoch. To train the CNN model for 10 epochs on the first participant's data took around 151.88 seconds and that time was 573.58 seconds for the LSTM model. That suggests that the CNN model can be trained for around 38 epochs within a time that can train the LSTM model for 10 epochs. Therefore, the CNN model was trained for 38 epochs. It took the CNN model 571.02 seconds to train on the first participant's data with 38 epochs, and 330.54 seconds on average per participant to finish training on 22 participants' data.

\begin{figure}[th]
     \centering
     \begin{subfigure}{0.46\textwidth}
         \centering
         \includegraphics[width=\columnwidth]{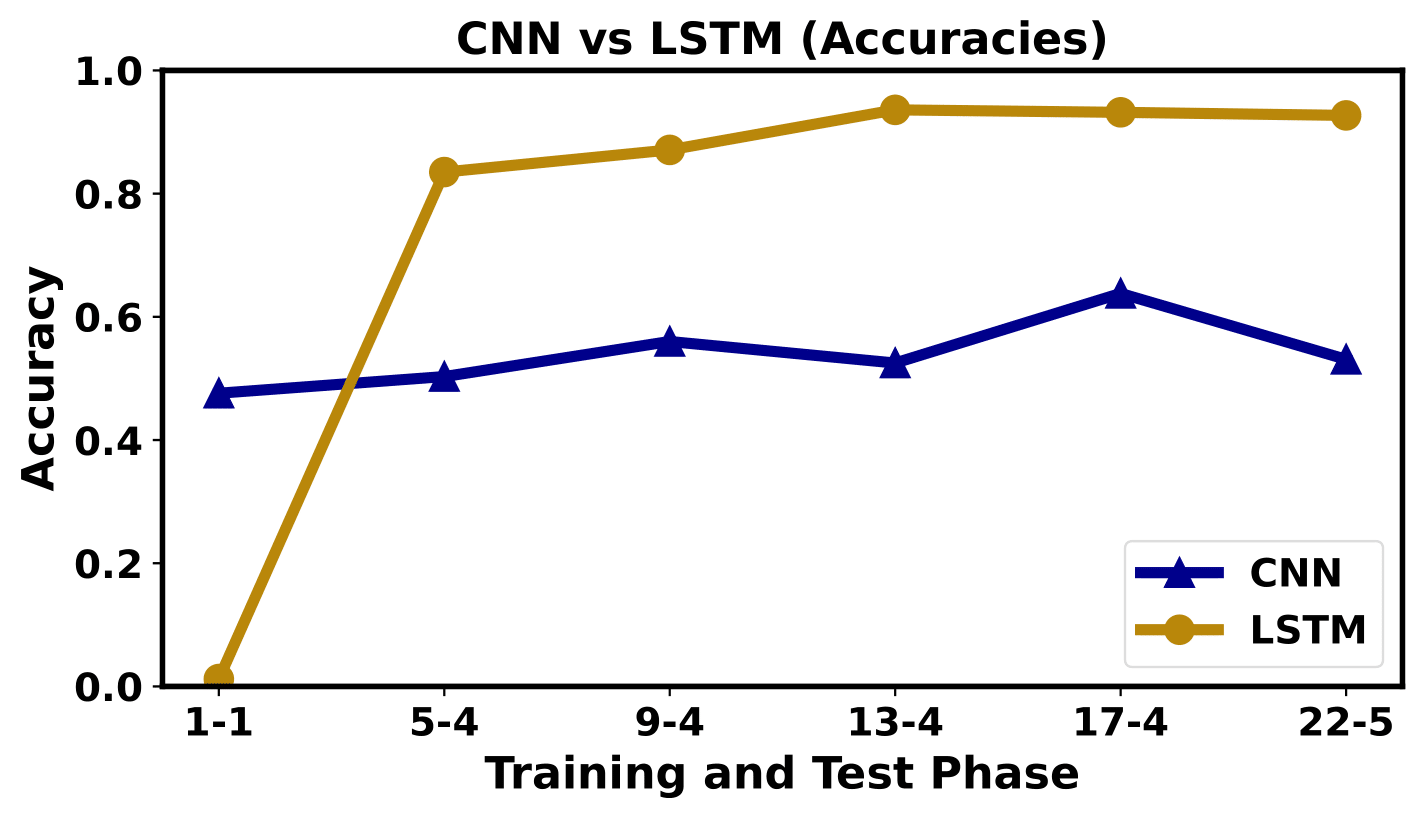}
         \caption{Accuracies}
     \end{subfigure}
     \begin{subfigure}{0.46\textwidth}
         \centering
         \includegraphics[width=\columnwidth]{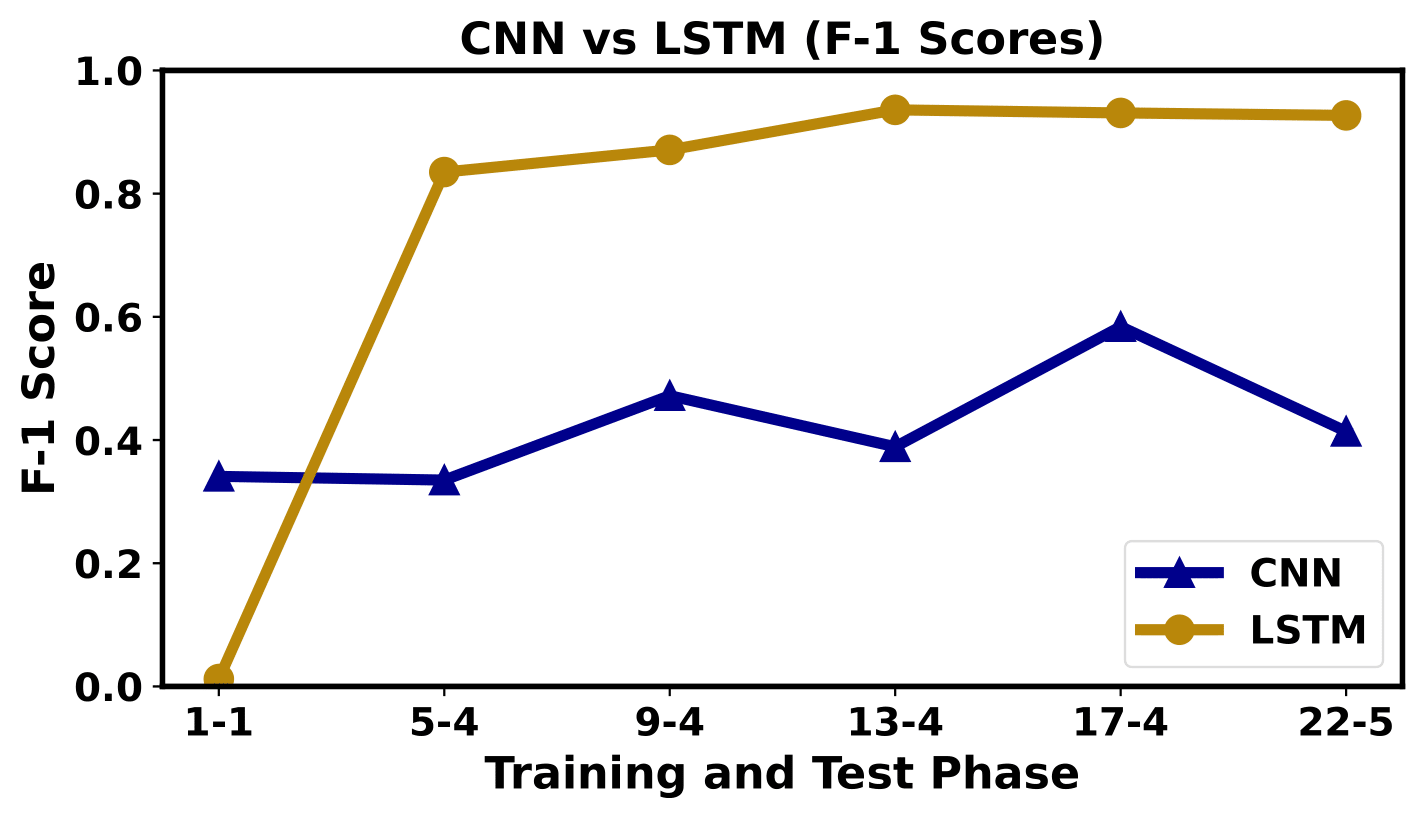}
         \caption{F-1 scores}
     \end{subfigure} 
\caption{Comparison of the CNN and LSTM models' accuracies and macro average F-1 scores in different training phases. The training phase labels, 1-1, 5-4, 9-4, 13-4, 17-4, and 22-5, indicate the numbers of training and test participants for those phases.}
\label{fig:cnn_v_lstm}
\end{figure}

\begin{table}[htbp]
    \centering
    \caption{Compute time needed with CPU for training the LSTM models. H = high-resolution, L = low-resolution, and K = future knowledge features.}
    \begin{tabular}{llllll}
        \hline
         \#Train  & Input & Feature & \#Param & Epochs & Train  \\
         users & & dim. & & &  hours\\
         \hline
         22  & H & 14 & 139 & 10 & 1.89 \\
         22 & H+K & 40 & 347 & 10 & 1.93 \\
         22  & H+L & 53 & 451 & 10 & 2.40 \\
         22  & H+L+K & 79 & 659 & 10 & 2.47 \\
         \hline
    \end{tabular}
    \label{tbl:compute_details}
\end{table}

It was found that the LSTM model provided better accuracies, precisions, recalls, and F-1 scores than the CNN model in the later training phases. The accuracies and F-1 scores of these two models are compared in \figref{cnn_v_lstm}. Each LSTM model was trained for 10 epochs for every experimental session. The time taken to complete the training is presented in \tblref{compute_details}. Although efforts were made so that no other user-initiated heavy-duty applications were running while training the models, compute time can depend on other processes or background services that might be running on the machine.

\subsection{Improving the Performance with Knowledge Features}
In order to find the importance of future knowledge, ablation studies on the features were performed. We train the models with a few different sets of features and present their results in \tblref{feature_ablation}. From the experiments, it can be noticed that among all four different sets of features, H+L+K (high-sampled+low-sampled+knowledge) usually performs better as we increase the size of the training data. In \tblref{feature_ablation2}, it is noticed that between the results for feature sets H and H+K, H+K consistently gets higher precision, recall, and F-1 scores than H alone.  when both models were trained with 22 participants' training data and tested on 6 participants' test data.
The situation of \eqnref{third_forecaster} and \eqnref{fourth_forecaster} is simulated in the rows with the H and H+K features. In \tblref{feature_ablation}, we see that when the model is trained with 22 participants' data, they get an F-1 score of 0.442 with only sensor data, but the F-1 score is 0.936 when future knowledge is added. This is an improvement of almost 112\%. This implies that by adding only a few knowledge parameters, a well-designed neural network can possibly forecast a medication event variable without even using the past values of that same variable. Another set of experiments was conducted to validate the hypothesis and similar findings were observed. In \tblref{feature_ablation2}, H+K features provided an accuracy as high as 0.936 along with all three other metrics, precision, recall, and F-1 score values greater than or equal to 0.936.

\begin{table}[!h]
    \centering
    \caption{Ablation experiments for training in 6 iterations.  Four LSTM models are independently trained for different sets of features: H= high-resolution, L= low-resolution, and K= future knowledge features. The p-value for the effect of K on F-1 scores is 0.00006 $<<$ standard threshold of 0.05.}
    \begin{tabular}{llllllll}
        \hline
         \#Train & \#Test & Features & Accuracy       & Precision      & Recall         & F-1            \\
users   & users  &          &                &                &                &                \\  \hline
1       & 1      & H        & 0.617          & 0.189          & 0.521          & 0.237          \\
1       & 1      & H+K      & 0.869          & 0.385          & 0.562          & 0.449          \\
1       & 1      & H+L      & 0.510          & 0.015          & 0.333          & 0.028          \\
1       & 1      & H+L+K    & \textbf{0.937} & \textbf{0.405} & \textbf{0.600} & \textbf{0.469} \\  \hline
5       & 4      & H        & 0.549          & 0.447          & 0.328          & 0.375          \\
5       & 4      & H+K      & 0.805          & 0.847          & 0.719          & 0.733          \\
5       & 4      & H+L      & 0.659          & 0.652          & 0.820          & 0.705          \\
5       & 4      & H+L+K    & \textbf{0.912} & \textbf{0.895} & \textbf{0.936} & \textbf{0.913} \\  \hline
9       & 4      & H        & 0.475          & 0.321          & 0.441          & 0.361          \\
9       & 4      & H+K      & \textbf{0.909} & \textbf{0.871} & 0.961          & \textbf{0.914} \\
9       & 4      & H+L      & 0.782          & 0.824          & 0.715          & 0.765          \\
9       & 4      & H+L+K    & 0.896          & 0.848          & \textbf{0.970} & 0.904          \\  \hline
13      & 4      & H        & 0.494          & 0.166          & 0.333          & 0.222          \\
13      & 4      & H+K      & 0.845          & 0.878          & 0.801          & 0.838          \\
13      & 4      & H+L      & 0.833          & 0.787          & 0.921          & 0.847          \\
13      & 4      & H+L+K    & \textbf{0.919} & \textbf{0.888} & \textbf{0.958} & \textbf{0.922} \\  \hline
17      & 4      & H        & 0.605          & 0.665          & 0.659          & 0.616          \\
17      & 4      & H+K      & \textbf{0.906} & \textbf{0.855}         & \textbf{0.977} & \textbf{0.912} \\
17      & 4      & H+L      & 0.834          & 0.794          & 0.899          & 0.842          \\
17      & 4      & H+L+K    & 0.894          & 0.849 & 0.959          & 0.901          \\  \hline
22      & 5      & H        & 0.516          & 0.455          & 0.526          & 0.442          \\
22      & 5      & H+K      & \textbf{0.932} & \textbf{0.888} & \textbf{0.989} & \textbf{0.936} \\
22      & 5      & H+L      & 0.829          & 0.798          & 0.903          & 0.837          \\
22      & 5      & H+L+K    & 0.898          & 0.867          & 0.946          & 0.903 \\
         \hline
    \end{tabular}
    \label{tbl:feature_ablation}
\end{table}

\subsection{Importance of Personalization}
\begin{figure}[!t]
     \centering
     \begin{subfigure}{0.48\textwidth}
         \centering
         \includegraphics[width=\columnwidth]{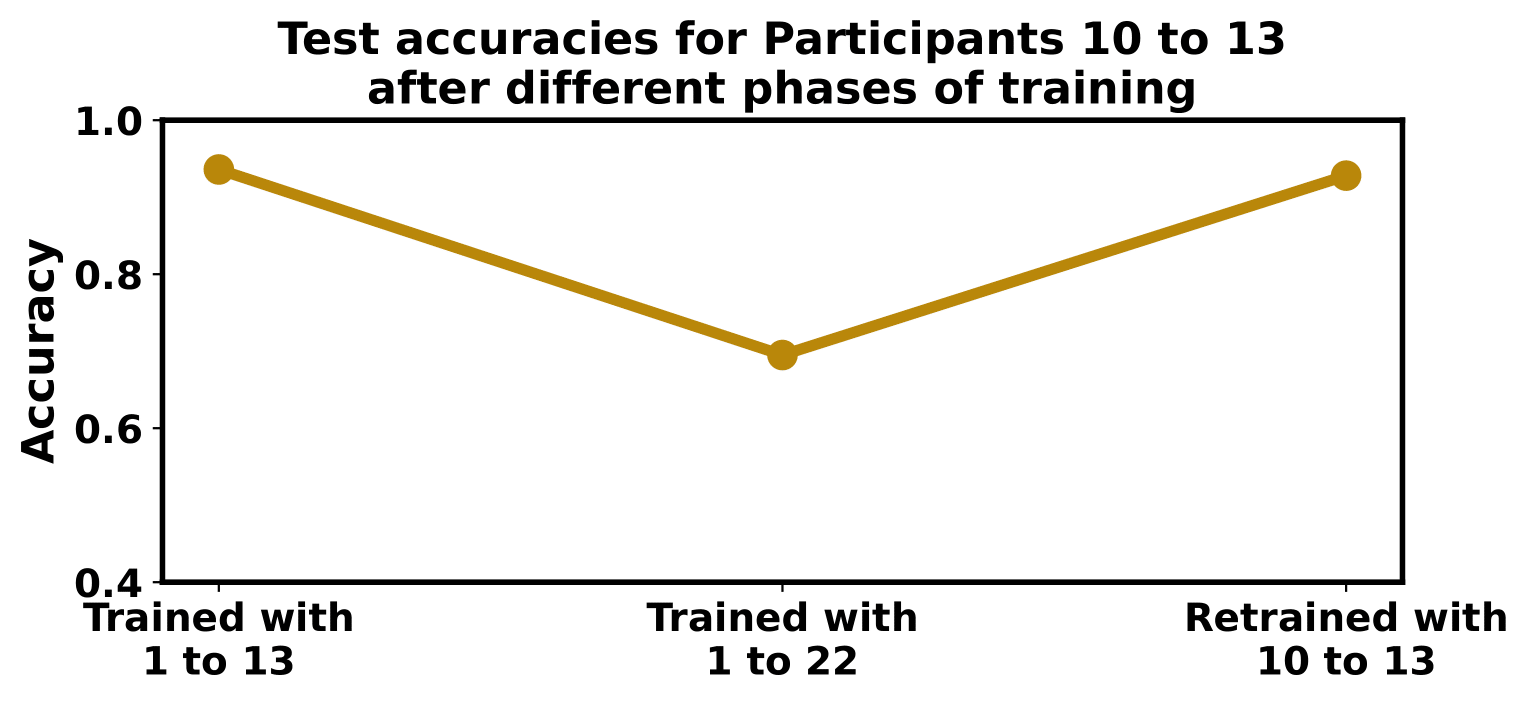}
         \caption{Accuracies}
     \end{subfigure}
     \begin{subfigure}{0.48\textwidth}
         \centering
         \includegraphics[width=\columnwidth]{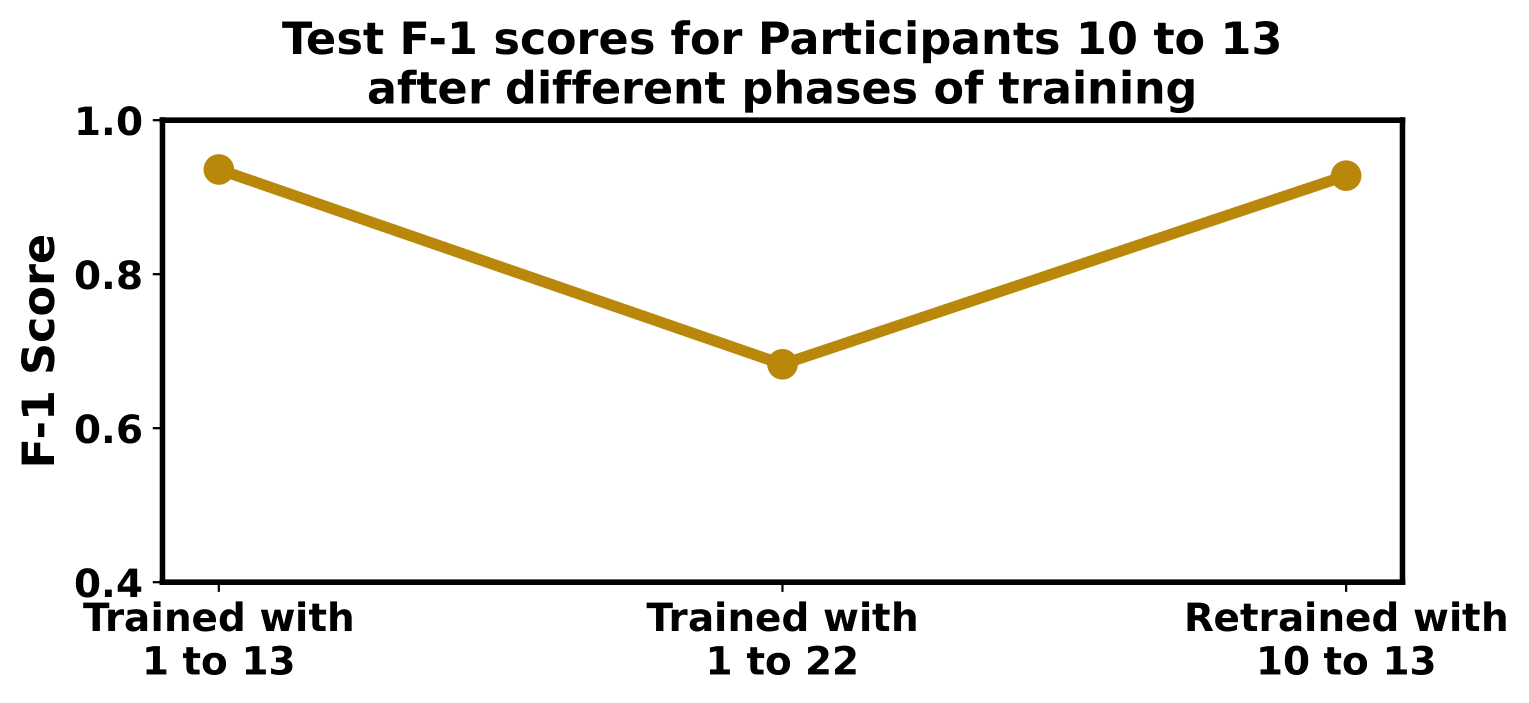}
         \caption{F-1 scores}
     \end{subfigure} 
\caption{The performance of the model drops for the test data of Participants 10 to 13 when the model is trained with 9 more additional participants. When the model is trained with the training data of Participants 10 to 13 again, the performance recovers on the test data of those participants.}
\label{fig:personalization}
\end{figure}

The performance of the ``fully'' trained LSTM model with sensor, event, and knowledge features can be the last row of \tblref{feature_ablation}. By the word, fully, we mean that the model was trained on all 22 participants' training data. The result reported in \tblref{feature_ablation} is on the last 5 participants' test data out of those 22 participants. Furthermore, this model was last trained at the 22-5 phase with Participants 18 to 22. In the fourth phase (10 to 13 phase), the participant IDs in the training data were 10 to 13. For a specific trial, when a fully-trained model is tested again on Participants 10 to 13, we noticed a drop in the F-1 score from 0.936 to 0.683. This happened because the fourth phase model forgot the ``best'' parameter values for the participants when it was trained with different participants in the fifth and sixth phases. When this model was trained again on the training data of Participants 10 to 13, we saw that the F-1 score improved from 0.683 to 0.928. The result of this experiment is depicted in \figref{personalization}. However, when the same experiment was repeated for Participants 14 to 17, the results were not similar. The three F-1 scores for the trial of this experiment on Participants 14 to 17 after Phase 5, after Phase 6, and after retraining on the data of Participants 14 to 17 are 0.931, 0.914, and 0.897 respectively. Nonetheless, these results suggest that personalization should be considered for participant behavioral action forecasting problems.

\subsection{Forecasting with Location and Future Knowledge Features}
In \tblref{feature_ablation_loc}, the forecasting accuracy and macro average precision, recall, and F-1 scores are presented for location and future knowledge features. The final two phases of test results suggest that location may be a very important factor in forecasting adherence when combined with future knowledge. If we compare the result with the results of \tblref{feature_ablation}, we notice that when the model has access to all sensors, the performance improves very quickly as we feed more data to the model. When it has only the location features, it needs more data to learn the patterns and to achieve an F-1 score close to 0.90.

\begin{table}[!t]
    \centering
    \caption{Ablation experiments for training in 5 iterations. Participants for each phase are chosen differently than the experiments of \tblref{feature_ablation}. H= high-resolution, L= Low-resolution, and K= future knowledge features. The p-value for the effect of K on F-1 scores is 0.001 $<<$ standard threshold of 0.05. Therefore, our results are statistically significant.}
    \begin{tabular}{llllllll}
        \hline
         \#Train & \#Test & Features & Accuracy & Precision & Recall & F-1  \\
          users &  users  &  &  & & &  \\
         \hline
         1 & 1  & H & 0.905 & 0.575 & 0.574 & 0.575 \\
         1 & 1  & H+K  & \textbf{0.952} & \textbf{0.577} & \textbf{0.599} & \textbf{0.588}\\
         \hline
         8 & 7  & H & 0.403 & 0.241 & 0.406 & 0.292\\
         8 & 7  & H+K & \textbf{0.879} & \textbf{0.879} & \textbf{0.879} & \textbf{0.879}\\
         \hline
         12 & 4  & H & 0.525 & 0.526 & 0.524 & 0.518  \\
         12 & 4  & H+K  & \textbf{0.880} & \textbf{0.880} & \textbf{0.880} & \textbf{0.880} \\
         \hline
         16 & 4  & H & 0.490 & 0.489 & 0.490 & 0.485 \\
         16 & 4  & H+K  & \textbf{0.936} & \textbf{0.943} & \textbf{0.937} & \textbf{0.936}\\
         \hline
         22 & 6  & H & 0.516 & 0.549 & 0.518 & 0.428\\
         22 & 6  & H+K & \textbf{0.922} & \textbf{0.932} & \textbf{0.922} & \textbf{0.921}\\
         \hline
    \end{tabular}
    \label{tbl:feature_ablation2}
\end{table}

\begin{table}[!b]
    \centering
    \caption{Performance of forecasting with location features. Two LSTM models are independently trained for different sets of features: Loc = location, and K= future knowledge features. The p-value for the effect of K on F-1 scores is 0.00013 $<<$ standard threshold of 0.05.}
    \begin{tabular}{llllllll}
        \hline
         \#Train & \#Test & Features & Accuracy       & Precision      & Recall         & F-1            \\
users   & users  &          &                &                &                &                \\ \hline
1       & 1      & Loc      & 0.500          & 0.181          & 0.667          & 0.250          \\
1       & 1      & Loc+K    & \textbf{0.543} & \textbf{0.329} & \textbf{1.000} & \textbf{0.402} \\ \hline
5       & 4      & Loc      & 0.501          & 0.168          & 0.333          & 0.223          \\
5       & 4      & Loc+K    & \textbf{0.756} & \textbf{0.578} & \textbf{0.602} & \textbf{0.589} \\ \hline
9       & 4      & Loc      & 0.500          & 0.166          & 0.333          & 0.221          \\
9       & 4      & Loc+K    & \textbf{0.889} & \textbf{0.872} & \textbf{0.911} & \textbf{0.888} \\ \hline
13      & 4      & Loc      & 0.503          & 0.336          & 0.667          & 0.446          \\
13      & 4      & Loc+K    & \textbf{0.906} & \textbf{0.858} & \textbf{0.978} & \textbf{0.913} \\ \hline
17      & 4      & Loc      & 0.502          & 0.168          & 0.333          & 0.223          \\
17      & 4      & Loc+K    & \textbf{0.876} & \textbf{0.847} & \textbf{0.921} & \textbf{0.879} \\ \hline
22      & 5      & Loc      & 0.501          & 0.334          & 0.667          & 0.445          \\
22      & 5      & Loc+K    & \textbf{0.925} & \textbf{0.870} & \textbf{1.000} & \textbf{0.930} \\
         \hline
    \end{tabular}
    \label{tbl:feature_ablation_loc}
\end{table}

\section{Discussion}
\label{sec:discussion}
\subsection{Clinical Significance and Novelty}
The clinical significance of this work lies in its potential to improve medication adherence in patients with heart failure (HF) by addressing a critical and often overlooked aspect of disease management: the forecasting of adherence based on temporal and contextual factors. HF affects millions of people in the U.S., with high morbidity and mortality rates, yet non-adherence to prescribed medications remains a common challenge that leads to poor outcomes. By leveraging temporal and behavioral data, including contextually relevant information such as time of day, location, and activities, this study seeks to identify patterns that may predict non-adherence, which can inform more effective, personalized interventions.

This work closes this gap in adherence forecasting by introducing a model that uses future contextual knowledge, which has not been thoroughly explored in current literature, to enhance prediction accuracy. Through the proposed CNN and LSTM-based approaches, this research provides a methodological advancement by addressing the challenges of sparse event forecasting and evaluating model performance on real-world healthcare data. Ultimately, this research has the potential to contribute to the development of tailored intervention systems, enabling healthcare providers to proactively address adherence barriers, thus improving patient outcomes and reducing healthcare costs associated with heart failure.

\subsection{Summary of Our Contribution and Findings}

This paper highlights the importance of medication adherence in chronic cardiovascular conditions, reviewing prior work and providing an overview of time-series forecasting using deep learning and classical ML algorithms, along with key challenges and solutions. We designed a forecasting system, AIMI, to solve the treatment adherence problem. To develop and evaluate the system in free-living settings, we conducted a user study with individuals with cardiovascular diseases. We then trained CNN and LSTM models to complete the system development.

Our LSTM-based models achieve an accuracy of 0.932 and an F-1 score of 0.936 on the test set. We found that the location and future knowledge features play an important role in improving performance. Moreover, in free-living settings, personalization of the trained models is also an important step to overcome the forecasting system's parameter forgetting challenge. When the model is trained with only one user’s data, the data distribution it has seen is very small and the model’s learning is somewhat random. Adding more data to the training set improves the performance. Moreover, when more features are added to the feature set, it requires more data to train the model properly. If we look at the results of the last few rows of \tblref{feature_ablation} when the model is ``fully-trained'', we see that the results are improving with the inclusion of K features.

\subsection{Limitations and Future Works}
The experiments of this paper show the importance of using future knowledge in the feature set for time-series forecasting. Nonetheless, the study has a few limitations. First of all, the dataset is highly skewed towards one class label. Medication events are sparse events and adherence varies highly across individuals. The number of representative samples for both classes in the eight-week study could not be ensured. Moreover, as the problem addresses hourly forecasting and the medication event usually takes place only once a day, the negative samples outnumber the positive samples by a factor of almost 24 in some cases. For this reason, the test set was balanced for a justified evaluation of the forecasting models.

A future direction of this work is the deployment of the forecasting model in a smartphone application where the model will track the sensor data and recent medication events and forecast whether a person may miss their medication. The smartphone app can then notify the user if the model forecasts they might miss their medication. A clinical trial may be performed to measure the efficacy of the system.

\section{Conclusion}
\label{sec:conclusion}
Forecasting sparse events is a challenging task, especially when the data is collected in an uncontrolled environment. This particular area is hardly explored in deep learning research. Moreover, the use of future knowledge is a relatively new concept that has not been applied in sparse events forecasting to the best of our knowledge. This paper aimed to contribute to this gap by evaluating CNN and LSTM models using high-sampled and low-sampled features, with and without future knowledge. The accuracies, and macro average precision, recall, and F-1 scores of the knowledge-guided models tend to outperform the knowledge-free models after the model is trained with a certain size of data. When only high-sampled features and future knowledge data are available, the improvement in the F-1 score using future knowledge improves by almost 112\%. 

This paper also addresses the importance of personalization in behavioral sparse event forecasting. It has been shown that a model may forget the best weight parameters for particular participants as it keeps training on additional participants' data. Hence, retraining with the target participants' data is important before making predictions in a deployed system. We invite more researchers to join this area so that the still unanswered questions can be answered with substantial theoretical and empirical evidence. We will share a public version of the medication adherence dataset in the near future after removing sensitive information. 

\section*{Acknowledgments}

This work was supported in part by the National Institutes of Health under grant R21NR015410 and by the National Science Foundation under grant CNS-2227002.  Any opinions, findings, conclusions, or recommendations expressed in this material are those of the authors and do not necessarily reflect the views of the funding organizations.

\bibliographystyle{elsarticle-harv} 
\bibliography{main}
\linenumbers

\newpage
\appendix
\section{Architecture of the CNN model}
\label{sec:app1}

Although the code for the full pipeline is shared separately, here we share the code for the CNN model for convenience. The CNN model used in the main paper contained a series of convolutional layers followed by a `flatten' layer and a dense layer.
\\

\begin{verbatim}
from tensorflow.keras.models import Model
from tensorflow.keras.layers import Flatten, Dense
from tensorflow.keras.layers import Conv2D
from tensorflow.keras.layers import Reshape, Input, Add, Activation
from tensorflow.keras import activations

def create_model_cnn(window_size, feature_size):    
    input_frame = Input(shape=(window_size, feature_size, 1))
    x = Conv2D(10, (60, 1), strides=(30,1), activation='relu')(input_frame)
    x = Conv2D(10, (1, 1), strides=(1,1), activation='relu')(x)
    x = Conv2D(10, (1, 1), strides=(1,1), activation='relu')(x)
    x = Conv2D(10, (1, 1), strides=(1,1), activation='relu')(x)
    x = Conv2D(10, (1, 1), strides=(1,1), activation='relu')(x)
    x = Conv2D(10, (1, 1), strides=(1,1), activation='relu')(x)
    x = Flatten()(x)
    x = Dense(1, activation='relu')(x)
    x2 = Reshape((window_size*feature_size*1,))(input_frame)
    x2 = Dense(1)(x2)
    z = Add()([x, x2])
    output = Activation(activations.sigmoid)(z)
    return Model(input_frame, output)
\end{verbatim}

\begin{figure}[h!]
     \centering
         \includegraphics[width=0.70\textwidth]{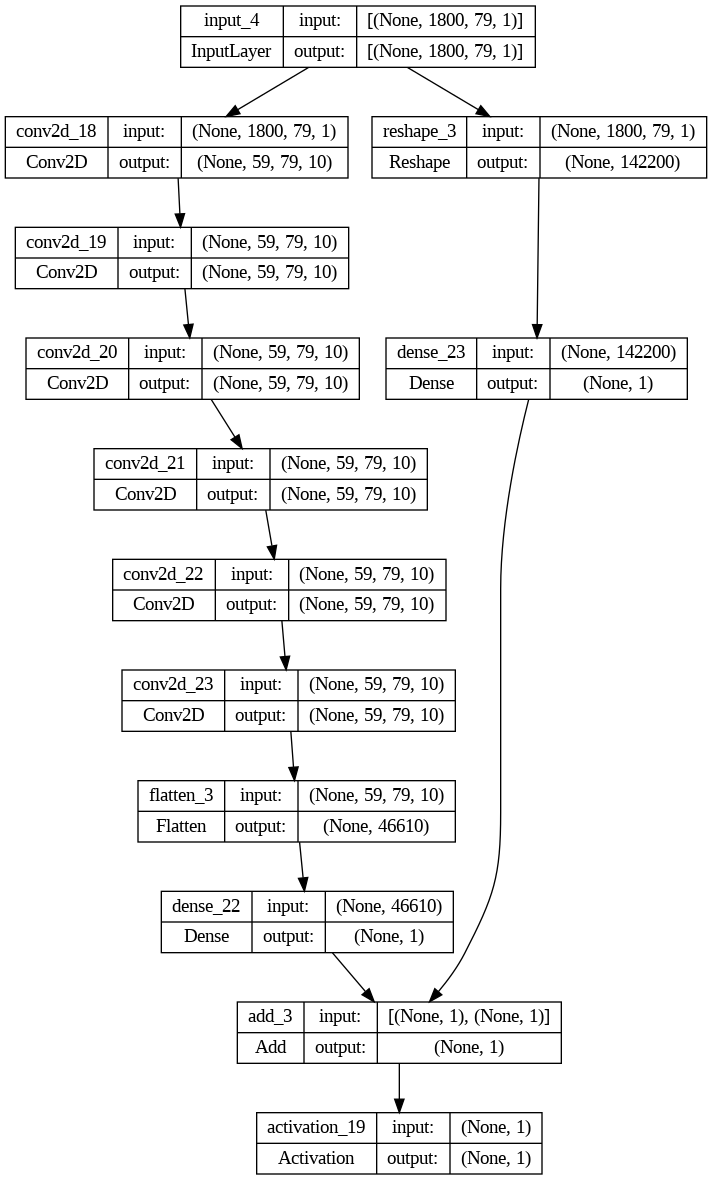}
\caption{The architecture of the CNN model.}
\label{fig:cnn_arch}
\end{figure}

 The model also included a skip connection from the input feature to the last hidden layer through a dense layer in between. A logical diagram of the architecture of the CNN model that uses all high-resolution, low-resolution, and knowledge features is presented in \figref{cnn_arch}. The CNN models for other feature sets have similar architecture except for the input shape of the first layer, which eventually affects the output and input shape of most layers.

\end{document}